\documentclass[12pt,a4paper]{article}

\usepackage[hmargin=1in,vmargin=1in]{geometry}
\usepackage{graphicx}
\usepackage{authblk}
\usepackage{titlesec}
\usepackage{caption}
\usepackage{fancyhdr}
\usepackage{url}
\usepackage[superscript,biblabel]{cite}
\usepackage{parskip} 
\usepackage{soul}
\usepackage{amsmath}

\usepackage{xcolor} 

\titleformat{\section}[block]{\normalfont\normalsize\bfseries}{\thesection.}{1em}{}
\titleformat{\subsection}[block]{\normalfont\normalsize\bfseries}{\thesection.}{1em}{}

\title{\Large Everything-Grasping (EG) Gripper: A Universal Gripper with Synergistic Suction–Grasping Capabilities for Cross-Scale and Cross-State Manipulation}

\author[1]{Jianshu Zhou$^\dagger$}
\author[2,3]{Jing Shu$^\dagger$}
\author[2]{Tianle Pan}
\author[3]{Puchen Zhu}
\author[4,5]{Jiajun An}
\author[3]{Huayu Zhang}
\author[3]{Junda Huang}
\author[5]{Upinder Kaur}
\author[3,4]{Xin Ma$^*$}
\author[1]{Masayoshi Tomizuka}

\affil[1]{Department of Mechanical Engineering, University of California, Berkeley.}
\affil[2]{Department of Biomedical Engineering, The Chinese University of~Hong~Kong.}
\affil[3]{Department of Mechanical and Automation Engineering, The Chinese University of Hong Kong.}
\affil[4]{Multi-scale Medical Robotics Center, Hong Kong, China.}
\affil[5]{Department of Agricultural and Biological Engineering, Purdue University.}

\date{}

\begin{document}
\maketitle

\let\thefootnote\relax
\footnotetext{$^\dagger$ Joint First Authors.}
\footnotetext{$^*$ Corresponding Author}

\vspace{-1em}

\vspace{1.2em}
\renewcommand{\abstractname}{\normalsize\bf Abstract}

\begin{abstract}
Grasping objects across vastly different sizes and physical states—including both solids and liquids—with a single robotic gripper remains a fundamental challenge in soft robotics. We present the Everything-Grasping (EG) Gripper, a soft end-effector that synergistically integrates distributed surface suction with internal granular jamming, enabling cross-scale and cross-state manipulation without requiring airtight sealing at the contact interface with target objects.The EG Gripper can handle objects with surface areas ranging from sub-millimeter scale 0.2 mm$^2$ (glass bead) to over 62,000 mm$^2$ (A4 sized paper and woven bag), enabling manipulation of objects nearly 3,500$\times$ smaller and 88$\times$ larger than its own contact area (approximated at 707 mm$^2$ for a 30 mm-diameter base). We further introduce a tactile sensing framework that combines liquid detection and pressure-based suction feedback, enabling real-time differentiation between solid and liquid targets. Guided by the actile-Inferred Grasping Mode Selection (TIGMS) algorithm, the gripper autonomously selects grasping modes based on distributed pressure and voltage signals. Experiments across diverse tasks—including underwater grasping, fragile object handling, and liquid capture—demonstrate robust and repeatable performance. To our knowledge, this is the first soft gripper to reliably grasp both solid and liquid objects across scales using a unified compliant architecture.
\end{abstract}


\newpage
\section*{I. Introduction}

From delicately picking up a fragile glass bottle to swiftly swiping away a spill, humans perform complex object manipulations with remarkable ease. Yet, despite decades of advancements, robots still struggle to match this adaptability and often falter when confronted
with diverse shapes, sizes, and physical states of real-world objects~\cite{billard2019trends}. Reliable robotic grasping typically relies on form closure, where the gripper encloses the object, or force closure, where friction resists disturbances~\cite{bicchi2000robotic, Prattichizzo2016}. These principles have guided rigid gripper design, supporting precision and robustness in industry~\cite{tai2016state}, but they remain unsuitable for fragile, irregular, or deformable objects.

To overcome these limitations, soft robotic grippers, built from compliant materials and mechanisms, have emerged as promising alternatives that offer adaptability and safe physical interaction~\cite{rus2015design, wang2018toward, shintake2018soft}. Soft grippers can be broadly categorized into contact-based and non-contact systems depending on their interaction modality~\cite{shintake2018soft, cianchetti2018biomedical, kim2013soft}. Non-contact approaches, which rely on external fields such as magnetic, electrostatic, or acoustic forces~\cite{kim2020untethered, yarali2022magneto}, show potential in microscale domains but remain limited in robustness and scalability for everyday use.

As a result, contact-based grasping dominates soft robotic manipulation due to its reliability and versatility in real-world settings~\cite{billard2019trends, Prattichizzo2016, rus2015design}. Functionally, contact-based soft grippers can be classified by their attachment mechanisms into two major categories: structure-based and adhesion-based designs.

Structure-based grippers achieve grasping through geometric enclosure, pinching, or adaptive deformation. Representative designs include fiber-reinforced actuators and Pneu-Nets that adapt to object contours~\cite{zhou2017soft, zhou2019soft, gu2021analytical, zhou2023antagonistic}; Fin Ray-inspired fingers for passive compliance~\cite{crooks2016fin, shin2021universal}; and soft–rigid hybrids that combine flexibility with structural integrity~\cite{zhou2024dexterous, zhou2020adaptive}. Other strategies include tentacle-like grippers for wrapping-based capture~\cite{sinatra2019ultragentle, yang2024multi}; jamming-based systems with tunable stiffness~\cite{nguyen2023granulartendon, amend2012positive, nguyen2023universally, li2017passive}; origami-inspired architectures for foldable, reconfigurable morphologies~\cite{li2019vacuum, cao2024design}; and smart-material-based designs for stimulus-responsive actuation.~\cite{kim2021fourD, cao2024design}.

Suction and adhesion-based soft grippers, by contrast, rely on surface forces for object attachment. These mechanisms include vacuum suction~\cite{koivikko2021magnetically, zhou2024design, zhou2022bioinspired}, dry adhesion inspired by gecko setae\cite{day2013microwedge, hawkes2016three}, wet adhesion for moist or underwater surfaces~\cite{nguyen2018soft, nguyen2019grasping}, electroadhesion via electrostatic attraction~\cite{cacucciolo2019delicate, gu2018soft}, and hybrid strategies that combine these principles~\cite{li2022bioinspired}.

Despite extensive progress, both structure-based and suction-based soft grippers exhibit intrinsic limitations. Structure-based grippers often struggle with flat, large, or highly deformable objects, while suction-based designs require airtight contact and tend to underperform on porous or irregular surfaces. Furthermore, many existing grippers are physically constrained by their own geometry, limiting their ability to manipulate objects larger than themselves. Although suction systems can lift larger items via localized negative pressure~\cite{bamotra2018fabrication, koivikko20213d, pham2019critically}, they typically perform best on smooth, sealed surfaces—conditions rarely found in unstructured environments such as households, where objects vary widely in size, shape, and texture.

To overcome the trade-offs between grasping and suction modes, hybrid-mode grippers have been proposed that integrate suction and structural grasping within a single system, aiming to combine the advantages and mitigate the limitations of single-mode designs. For instance, Liu et al.\cite{liu2023hybrid} and Wang et~\cite{wang2020dual}integrated suction with soft fingers, Wu et al.\cite{wu2022glowing} combined suction with origami-based morphologies, and Washio et al.~\cite{washio2022design} introduced a shape-morphing suction cup featuring three grasping modes. While these designs improve cross-scale grasping, they typically rely on discrete actuation modes and bulky structures. Suction is often treated as a secondary function using off-the-shelf components, which limits the ability to grasp objects smaller than the gripper itself and reduces scalability in real-world scenarios.

To extend the cross-scale grasping capability of hybrid-mode grippers, particularly for manipulating objects smaller than the gripper itself, we previously developed the Transformable Soft Gripper (TSG).~\cite{pan2024transformable}. The TSG unified suction and jamming based grasping within a single structure, enabling amphibious and cross-scale object handling across a 2–200 mm range. However, the system still required explicit mode switching (around 5 seconds), distinct control strategies for different object sizes, and visual feedback for reliable operation. Most critically, the TSG lacked the ability to handle liquids, limiting its applicability in cross-state manipulation tasks. Furthermore, due to the absence of integrated tactile and liquid sensing capabilities, unlike recent works on tactile-integrated grasping~\cite{wang2018toward, chathuranga2013biomimetic, qi2022sea}, the TSG could not autonomously infer object properties or adaptively select grasping modes, thereby limiting its applicability in unstructured or sensor-limited environments.

We present the Everything-Grasping (EG) Gripper, named for its versatility in manipulating a wide range of object types, including both solids and liquids, across extreme scales and diverse environments. To the best of our knowledge, it is the first soft robotic system to achieve both cross-scale and cross-state manipulation through the synergistic integration of suction and grasping. The EG Gripper operates robustly in air, on liquid surfaces, and underwater, enabling seamless multimodal handling across varied conditions. The EG gripper combines surface-mounted distributed suction elements (DSEs) with an internal granular jamming membrane, enabling a single grasping mode—surface-conformal adhesion—that requires no airtight sealing and remains robust across diverse material properties. It stably manipulates objects with surface areas from 0.2 mm² (e.g., glass bead) to over 62,000 mm² (e.g., A4 sheet or woven bag). As shown in Fig. 1, the EG gripper outperforms suction-focused designs in small-scale handling~\cite{washio2022design} and exceeds grasping-focused solutions in large-scale manipulation~\cite{cui2021enhancing}, offering superior cross-scale performance compared to prior hybrid suction–grasping systems~\cite{li2022bioinspired, wang2020dual, pan2024transformable}. It achieves a maximum grasping force of 6.14 N and uses a single grasping mode for all object sizes, simplifying control and reducing reliance on external vision. Guided by the dedicated Tactile-Inferred Grasping Mode Selection (TIGMS) algorithm, integrated pressure sensing and liquid detection modules provide vision-free, real-time feedback for autonomous differentiation between solid and liquid targets, as well as object size estimation and grasping modes selection. Extensive experimental validation—including underwater grasping, fragile object handling, and floating liquid capture—demonstrates the system’s robustness, repeatability, and adaptability across diverse manipulation scenarios.

The main contributions of this work are as follows:

\begin{enumerate}

    \item We propose the Everything-Grasping (EG) gripper, a unified soft robotic design that synergistically integrates surface-distributed suction and jamming. This is one of the first gripper enables reliable cross-scale and cross-state manipulation without mode switching, airtight sealing, or material-specific tuning.
    
    \item We develop a tactile sensing framework combining pressure feedback and liquid detection, and introduce the TIGMS algorithm to autonomously infer object state and adapt grasping strategies without visual input.
    
   \item We validate the EG gripper across diverse real-world scenarios—including submerged, humid, and floating conditions—demonstrating reliable manipulation of objects ranging from 0.2\,mm\textsuperscript{2} to 62{,}000\,mm\textsuperscript{2}. This corresponds to targets nearly 3{,}500 times smaller and 88 times larger than the gripper’s contact base (approximated at 707\,mm\textsuperscript{2} for a 30\,mm-diameter footprint), all achieved using a compact, single-mode architecture.

\end{enumerate}

\begin{figure}[t]
\centering
\includegraphics[width=13cm]{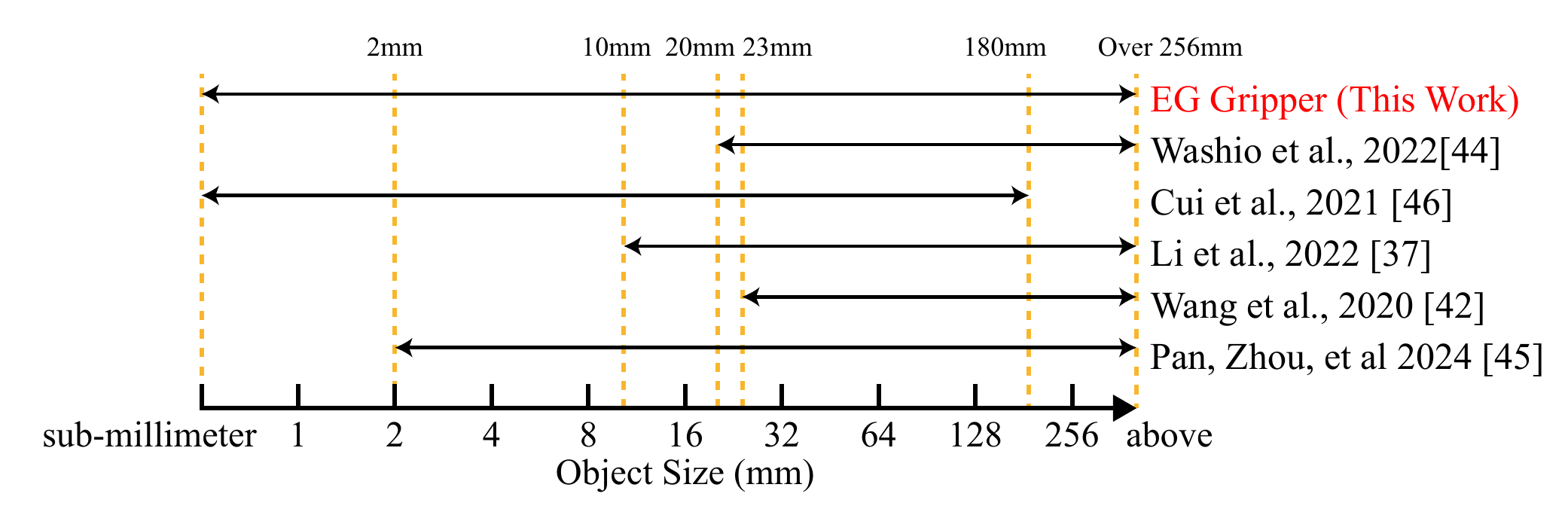}
\caption{Grasping range of the EG gripper compared to existing cross-scale grasping solutions.}
\label{Figure_intro}
\end{figure}

\section*{II. Design and Fabrication of the EG Gripper}

\subsection*{A. Structural Design and Working Principle}

The overall design of the EG gripper is shown in Fig.~2(a). The EG gripper consists of a silicone membrane, expandable polystyrene (EPS) particles, distributed suction elements (DSEs), pressure sensors, liquid detection modules, sealing caps, an adapter for the robotic arm, and pneumatic control systems including valves and micro vacuum pumps.

The fabrication process of the EG gripper is shown in Section A of the Supplementary Information. The silicone membrane is hemispherical, with a diameter of 30 mm and a thickness of 0.8 mm. It serves as the primary contact interface with grasped objects, and its base contact area is approximately 707 mm². The EPS particles, with a diameter of approximately 0.5~mm, are filled into the sealed silicone membrane to enable granular jamming by applying negative pressure.

There are nineteen DSEs in the EG gripper, each consisting of a bowl-shaped suction cup and a capillary. The capillaries extend through the entire EG gripper and connect the suction cup to the pressure sensors, liquid detection modules, and pneumatic servo systems at the distal end. The bowl-shaped suction cups are radially distributed on the surface of the silicone membrane. The radius of each suction cup is 2.5~mm. This bowl-shaped design increases the suction cup’s effective area when adhering to grasped objects.

Each DSE is connected to an individual micro vacuum pump and valves, along with a pressure sensor, which enables monitoring of the engagement states of the DSEs and independent control of the airflow. The liquid detection module for each DSE enables self-state sensing of the EG gripper during liquid capture.

In our design, the EG gripper employs a synergistic integration of micro-adhesion via DSEs and granular jamming, enabling cross-scale grasping and adaptability to complex object geometries, without the need for explicit mode switching, as both mechanisms operate concurrently. Fig.~2(b) shows schematics of object grasping with the EG gripper, illustrating both cross-scale and cross-state grasping. Nearby insets showcase the gripper’s bottom view, highlighting engaged (orange) and disengaged (gray) DSEs.

\begin{figure*}[t]
\centering
\includegraphics[width=0.8\textwidth]{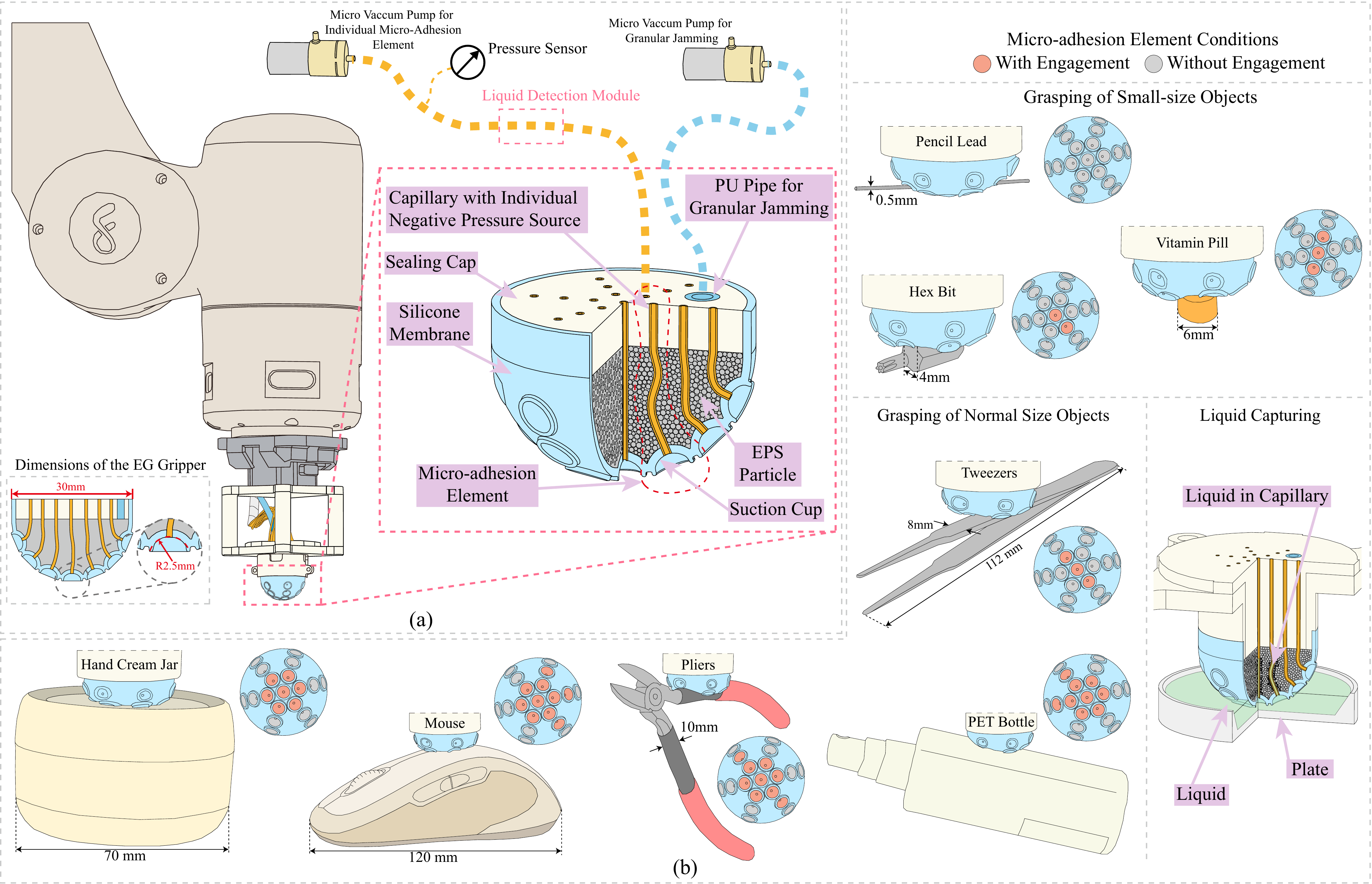}
\caption{Overall design of the proprioceptive Everything-Grasping (EG) gripper and its cross-scale and cross-state manipulation capabilities, demonstrating adaptability to complex object geometries.}
\label{Overall_Design}
\end{figure*}

\begin{figure*}[ht]
\centering
\includegraphics[width=0.8\textwidth]{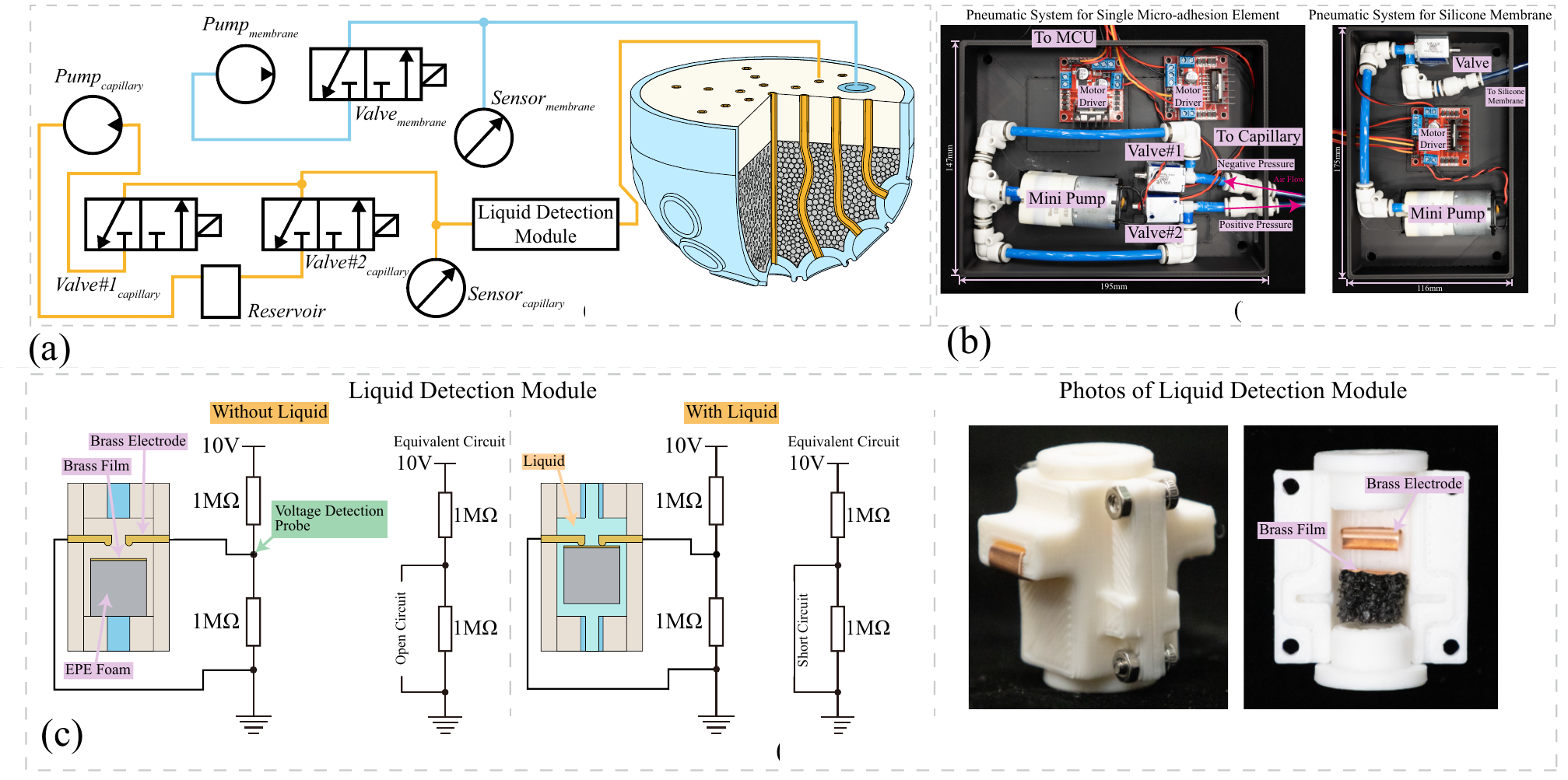}
\caption{Pneumatic control system and liquid detection module of the EG gripper.
(a) Schematic of the overall pneumatic control system integrating suction and jamming control.
(b) Detailed setup for a single micro-adhesion element and the silicone membrane interface.
(c) Liquid detection module for conductive liquids, featuring a vertically placed floating foam with a brass film that closes the circuit upon contact with the sidewall electrodes when buoyant force exceeds gravity.}
\label{pneumatic_circuit}
\end{figure*}

\subsection*{B. Pneumatic Control and Sensing Methods for the EG Gripper}

The sketch diagram for the pneumatic control system of the EG gripper is shown in Fig.~3(a). 
The blue lines represent the pneumatic circuit for the silicone membrane, and the yellow lines represent the pneumatic circuit for the DSE. 
In the pneumatic control system for the silicone membrane, there is a micro vacuum pump (SC3802PM-A, Skoocom, China) for generating negative pressure, and a two-position three-way solenoid valve (0526T, Fspump, China) for controlling the airflow. 
A pressure sensor (XGZP6869A, CFsensor Ltd, China) is used to monitor the pressure inside the silicone membrane. 
The pump, valve, and pressure sensor are placed at the distal end, and they are connected to the silicone membrane via a PU tube.

In the pneumatic control system for the DSE, there is a micro vacuum pump, two two-position three-way solenoid valves, a pressure sensor, and a liquid detection module. 
The two valves are connected to the inlet and outlet of the micro vacuum pump, respectively, to provide negative and positive pressure to the pneumatic circuit using a single pump. 
The photos of the pneumatic control systems are shown in Fig.~3(b). 
The valves and pumps are controlled by an MCU (Arduino Mega2560 R3, Arduino Srl, Italy) through motor driver modules (L298N, STMicroelectronics, Switzerland). 
The sensing signals from the pressure sensors are also read by the MCU through ADC modules (ADS1115, Adafruit, USA).

The photos and sketches of the liquid detection modules are shown in Fig.~3(c), together with their working mechanisms. 
The liquid detection module is placed vertically and utilizes the buoyancy of the liquid to detect its presence. 
There is a pair of brass electrodes placed in a rigid tube. 
The electrodes are connected to a circuit with a voltage detection probe. 
A gap with a width of 2~mm is designed between the two electrodes. 
Beneath the gap, there is a cylindrical expanded polyethylene (EPE) foam with a diameter of 9~mm. 
The upper surface of the EPE foam is bonded with a thin brass layer with a thickness of 0.1~mm. 
When there is liquid passing through the liquid detection module, the EPE foam will be lifted by buoyancy and come into contact with the electrodes, completing the circuit and causing a voltage change on the voltage detection probe. 
When there is no liquid in the liquid detection module, the EPE foam disconnects from the electrodes due to its weight.

Control algorithms are developed for the EG gripper to independently control the working states of each DSE. 
Before the contact between the EG gripper and the grasping object, the micro vacuum pumps in the pneumatic control system for the DSEs operate at 30\% of their maximum capacity, with pressure sensors continuously monitoring pressure fluctuations. 
Upon the EG gripper contacting the grasping object, the resulting pressure decrease is used to identify the number of engaged DSEs. 
The pumps associated with the engaged DSEs are then activated at full capacity, while those associated with the disengaged DSEs are deactivated. When the EG gripper is capturing liquid and the liquid detection module detects the liquid, the pump will be deactivated and the valve connected to the inlet of the pump will be turned off. 
When the EG gripper is injecting the liquid, the pump will be activated and the valve connected to the outlet of the pump will be turned on.

\begin{figure}[t]
\centering
\includegraphics[width=0.8\linewidth]{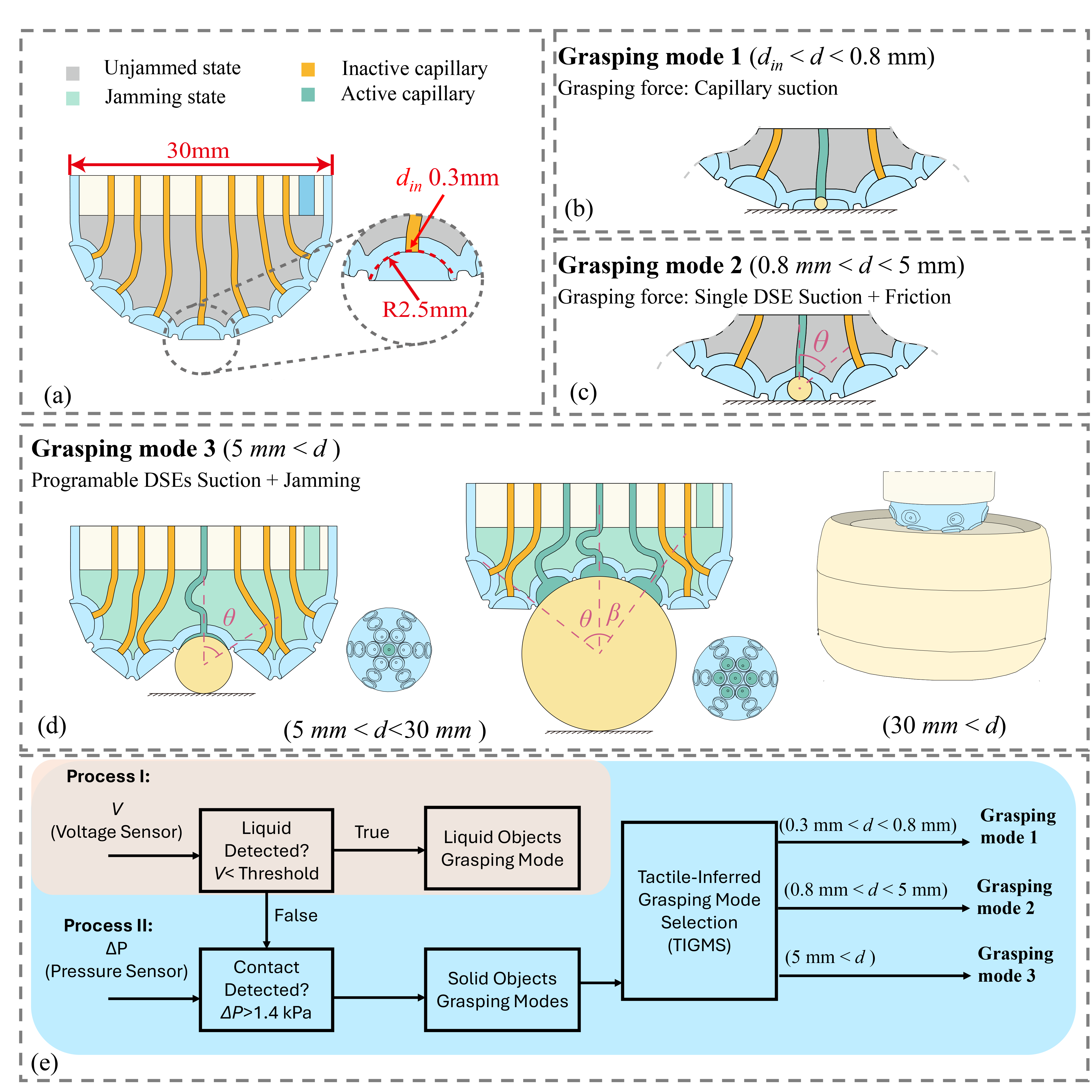}
\caption{Grasping modes and Tactile-Inferred
Grasping Mode
Selection (TIGMS) logic of the EG gripper. 
(a) Gripper structure with suction cup radius \( R_1 = 2.5\,\text{mm} \) and capillary diameter \( d_{\text{in}} = 0.3\,\text{mm} \). 
(b–d) Three synthetic solid-object grasping modes based on the inferred object dimension $d$.: 
Mode 1 (\(0.3 < d < 0.8\,\text{mm}\)) – capillary-only suction; 
Mode 2 (\(0.8 < d < 5\,\text{mm}\)) – suction with friction and jamming; 
Mode 3 (\(d > 5\,\text{mm}\)) – multi-DSE suction with shell jamming. 
(e) Logic illustration of Tactile-Inferred Grasping Mode Selection (TIGMS) using $\Delta P$ and $V$.
}
\label{figure_working_mechanism}
\end{figure}

\section*{III. Working Principle of the EG Gripper}
The principle underlying the EG gripper's ability to grasp cross-scale and cross-state objects is illustrated in Fig.~4. 
Each DSE generates suction force under applied negative pressure, which concurrently increases the normal contact force between the soft silicone membrane and the object's surface. 
When granular jamming is activated, the shell stiffens substantially, enhancing the frictional interface and enabling distributed force transmission. 
The suction and jamming mechanisms work synergistically: suction provides initial attachment and preload, while jamming reinforces the overall grasp by stabilizing the membrane structure. 
Without jamming, the membrane remains highly compliant, allowing the DSEs to passively adapt to diverse object geometries.

Depending on the size and contact condition of the target object, the grasping interaction involves capillary suction, suction-cup engagement, frictional locking, and distributed jamming, which are coordinated to adaptively generate grasping force. 
We define the effective suction cup diameter as $d_m = 5$\,mm. 
The following analysis presents analytical models for the EG gripper's grasping behavior, derived from physical interactions between suction and jamming under different object dimensions.

\subsection*{A. Analytical Model for the EG Gripper in Grasping Small-sized Objects ($d < d_m$)}

Two modes are considered in this scenario. In the first case, the object diameter is larger than the capillary diameter (0.3\,mm) but smaller than the maximum effective sealing diameter ($\sim$0.8\,mm), enabling the object to be grasped solely by capillary suction without engaging the suction cup surface as illustrated in Fig. 4(b). The threshold of 0.8\,mm was empirically determined using spherical beads with 0.1\,mm increments: objects up to 0.8\,mm could be reliably sealed by the capillary tip alone, whereas larger objects triggered peripheral contact with the suction cup surface. Geometrically, this threshold corresponds to the projected sealing diameter from the suction cup curvature $R_1 = 2.5$\,mm, following the relation $d = 2R_1 \sin \theta$ with $\theta \approx 9.2^\circ$, denoting the maximal contact angle between the cup surface and the object. 

In the second case, the object diameter ranges from approximately 0.8\,mm to 5\,mm (the effective suction cup diameter), such that the object makes partial contact with the suction cup surface as illustrated in Fig. 4(c). This contact leads to additional suction and frictional engagement, further enhanced by shell stiffening through granular jamming. The upper and lower portions of Fig.~4(b) illustrate these two grasping cases, respectively.

In the first case, the grasping objects only contact the end of the capillary, and the grasping force $F_g$ is primarily attributed to the suction force $F_s$ generated by the negative pressure within the capillary. 
Therefore, we have:
\begin{equation}
F_g = F_s = P_c A_0 \quad (d_{\text{in}} < d < 0.8~\text{mm})
\end{equation}
where $A_0$ is the effective cross-sectional area of the capillary, which is influenced by the selection of the capillary inner diameter $d_{\text{in}}$. 
In this case, $d_{\text{in}} = 0.3~\text{mm}$.

In the second case, the diameter of the grasping object $d$ ranges between 0.8 mm and 5 mm. 
The $F_g$ of the EG gripper has two components. 
The first component is the frictional force $F_f$, which is generated between the suction cup and the grasping object when granular jamming is applied. 
Following the analysis~\cite{brown2010universal}, $F_f$ can be analyzed based on a circular slip on the suction cup with a central angle $\delta\theta$. 
The width of the circular slip is $R_1\delta\theta$, where $R_1$ is the radius of the grasping object. 
The area $A_1$ of the slip is given by $A_1 = 2\pi R_1 \sin\theta \times R_1 \delta\theta$. 
The normal force $F_N$ between the suction cup and the grasping object can be expressed as $F_N = \sigma_0 A_1$, where $\sigma_0$ represents the normal stress. 
Finally, we obtain the following expression for $F_f$~\cite{brown2010universal}:
\begin{equation}
F_f = \int_0^{\theta_c} (\mu F_N \sin\theta - F_N \cos\theta) \, d\theta
\end{equation}

The second component of $F_g$ is $F_s$. 
Compared with the suction force represented in Eq.~1, the effective area increases from $A_0$ to $A_0^* = \pi (R_1 \sin\theta)^2$. 
Therefore, the overall grasping force $F_g$ in this case can be expressed as:
\begin{equation}
F_g = F_f + F_s = \int_0^{\theta_c} (\mu F_N \sin\theta - F_N \cos\theta) \, d\theta + P_c \pi (R_1 \sin\theta_c)^2 \quad (0.8~\text{mm} \le d < 5~\text{mm})
\end{equation}

\subsubsection*{B. Analytical Model for the EG Gripper in Grasping Large-sized Objects ($d \geq d_m$)}

When $d > 5$\,mm, as illustrated in Fig.~4(d), one or more DSEs will fully engage with the grasping object. For spherical objects, the number of engaged DSEs $n$ can be 1, 7, or 13, depending on the object's size. In this regime, the grasping force results from the combined effect of distributed suction and shell stiffening through granular jamming.

\noindent When $n = 1$, we have
\begin{equation}
    F_s = P_c \pi (R_1 \sin \theta_c^*)^2
\end{equation}
where $\theta_c^*$ is half of the central angle corresponding to the DSE, and it can be calculated as $\theta_c^* = \frac{1}{2}(L_A / R_1)$ (as shown in Fig.~4(c)).

\noindent When $n = 7$ (as shown in Fig.~4(d)), we have
\begin{equation}
    F_s = P_c A_0^* + 6 P_c A_0^* \cos \beta
\end{equation}
where $A_0^*$ represents the effective area under vacuum in the sucker, which could be calculated using $A_0^* = \pi (R_1 \sin \theta_c^*)^2$.

\noindent There will be more DSEs for larger grasping objects, i.e., $n = 13$. We have
\begin{equation}
    F_s = P_c A_0^* + 6 P_c A_0^* \cos \beta + 6 P_c A_0^* \cos (2\beta)        \quad (5~\text{mm} < d)
\end{equation}

For the frictional force $F_f$, which is another component of the grasping force $F_g$ generated by the EG gripper when grasping normal-sized objects, its expression is identical to that in Eq.~2. In this scenario, the contact angle $\theta$ is defined between the silicone membrane and the object being grasped. Finally, the $F_g$ generated by the EG gripper when grasping normal-sized objects is the sum of $F_s$ and $F_f$.

Based on the analytical models derived in Section~III (Eqs.~1–6), we use two tactile sensing channels: the pressure signal ($\Delta P$) from the distributed DSE array and the voltage signal ($V$) from the liquid detection module. Here, $\Delta P$ reflects suction engagement and object contact at each DSE, and is considered significant when exceeding a threshold (e.g., 1.4\,kPa). The voltage signal $V$ decreases upon contact with conductive liquids, enabling liquid identification (e.g., $V < 1.1$\,V). Using these signals, we implement a rule-based selection strategy to determine the appropriate grasping mode. This logic is formalized in Fig.~4(e) and Algorithm I, enabling the EG gripper to autonomously infer object size and state, and to select the corresponding grasping strategy without external sensing or mode switching.

\textbf{Algorithm I: Tactile-Inferred Grasping Mode Selection (TIGMS)}

\textbf{Input:} Pressure response $\Delta P_i$ from each DSE $i = 1\ldots N$, voltage signal $V$  
\textbf{Output:} Grasping Mode $\in \{$Mode 1, Mode 2, Mode 3, Liquid$\}$

\begin{enumerate}
  \item \textbf{If} $V < V_\text{threshold}$:  
  \hspace{1.5em} \textbf{Return} Liquid Mode (trigger spray or release)

  \item \textbf{Else if} only capillary $\Delta P$ is detected, and no DSE suction:  
  \hspace{1.5em} \textbf{Return} Mode 1 (Capillary-only suction)

  \item \textbf{Else if} 1–3 neighboring DSEs detect $\Delta P$:  
  \hspace{1.5em} \textbf{Return} Mode 2 (Suction + partial jamming)

  \item \textbf{Else if} more than 3 DSEs detect $\Delta P$:  
  \hspace{1.5em} \textbf{Return} Mode 3 (Multi-DSE suction + shell jamming)
\end{enumerate}

\begin{figure*}[t]
\centering
\includegraphics[width=0.8\textwidth]{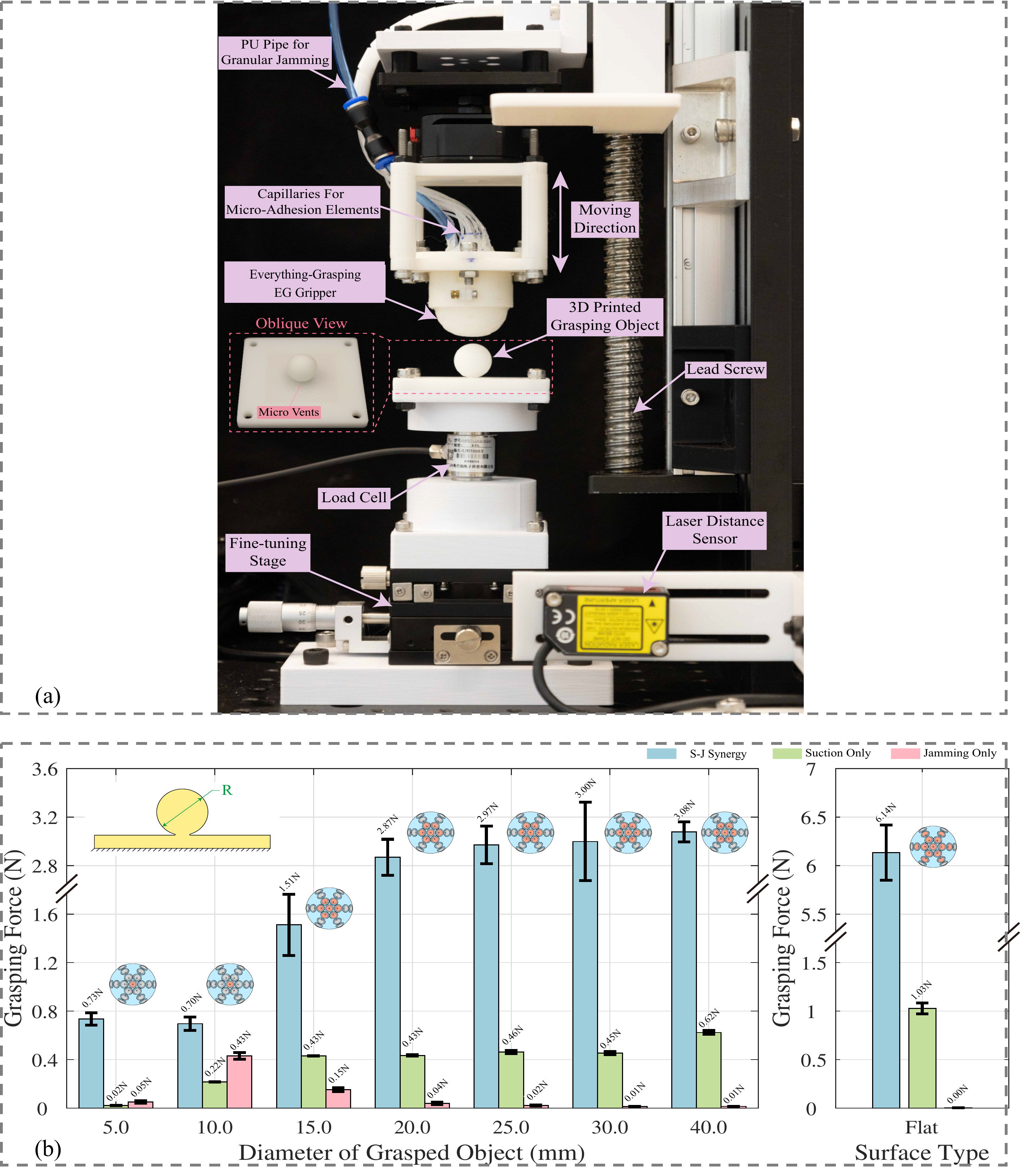}
\caption{Grasping force evaluation of the EG gripper. 
(a) Experimental setup. 
(b) Measured maximum grasping forces for spherical (left) and planar (right) objects. 
Synergistic, suction-only, and jamming-only modes are represented in blue, green, and pink, respectively.}
\label{figure_experiment_setup}
\end{figure*}

\section*{IV. EXPERIMENTAL RESULTS}

\subsection*{A. Grasping Force of the EG Gripper}

This section shows the force testing results of the EG gripper. The experimental setup is shown in Fig.~5(a). A force testing platform was built, which contains a vertically translatable stage driven by a stepper motor and lead screw, a load cell (SBT671, Simbatouch, China), and 3D-printed grasping objects. There are two shapes of grasping objects: spherical and flat-topped. The sizes of the spherical grasping objects range from 5~mm to 40~mm in diameter to evaluate the effect of object size on the grasping force of the EG gripper. For each grasping object, the grasping force is evaluated under three working modes of the EG gripper for comparison: synergistic mode, suction mode, and jamming mode. In synergistic mode, the DSEs and granular jamming work synergistically. In contrast, in suction mode and jamming mode, the DSEs and granular jamming work solely, respectively. Each force test was repeated three times.

The experimental results of the grasping force tests for the EG gripper are shown in Fig.~5(b). The bar graphs with different colors are used to represent the magnitude of the grasping force. The grasping forces generated by the EG gripper in synergistic mode, suction mode, and jamming mode are represented in blue, green, and pink, respectively. For each grasping object, the sketches of the active DSEs are illustrated alongside the corresponding bar graph. For spherical objects, the maximum grasping force of the EG gripper in synergistic mode increases with object diameter. When the EG gripper grasps a flat-topped object, it reaches the maximum grasping force (an average of 6.14~N). A substantial increase in grasping force is observed when the diameter of the grasping objects increases from 10~mm to 15~mm, and again when transitioning from the 40~mm diameter sphere to the flat-topped object, which correlates with a corresponding increase in the number of engaged DSEs. The observation aligns with the trends described in Eq.~4 to Eq.~6, which indicates that grasping force increases with both the number of engaged DSEs and a decrease in the angle $\beta$ (the angle between the DSE’s central axis and the grasping object surface). Notably, the synergistic action of DSEs and granular jamming resulted in a significantly increased grasping force that exceeds the sum of their individual contributions. The synergistic effect is particularly pronounced for grasping objects with diameters greater than 15~mm.

\subsection*{B. Engagement Detection and Liquid Ingress Monitoring of DSEs}

The experimental results of the engagement detection and liquid ingress monitoring of DSEs are shown in Fig.~6. Fig.~6(a) demonstrates the pressure changes when the silicone membrane of the EG gripper contacts the grasping objects. The micro vacuum pump operates at 30\% capacity at this moment. For the pressure of the engaged DSEs, it decreases by 1.42~kPa after 1~s and by 2.78~kPa after 5~s. The pressure change enables identification of engaged DSEs, triggering full-capacity operation of their associated micro vacuum pumps. Fig.~6(b) shows the voltage change detected by the liquid detection module at the moment of liquid ingress. The liquid ingress results in a rapid voltage drop from 5~V to 0.87~V within 4~ms. The significant voltage change allows the system to quickly identify DSEs that have captured liquid, enabling immediate pump deactivation.

\begin{figure}[t]
\centering
\includegraphics[width=0.7\linewidth]{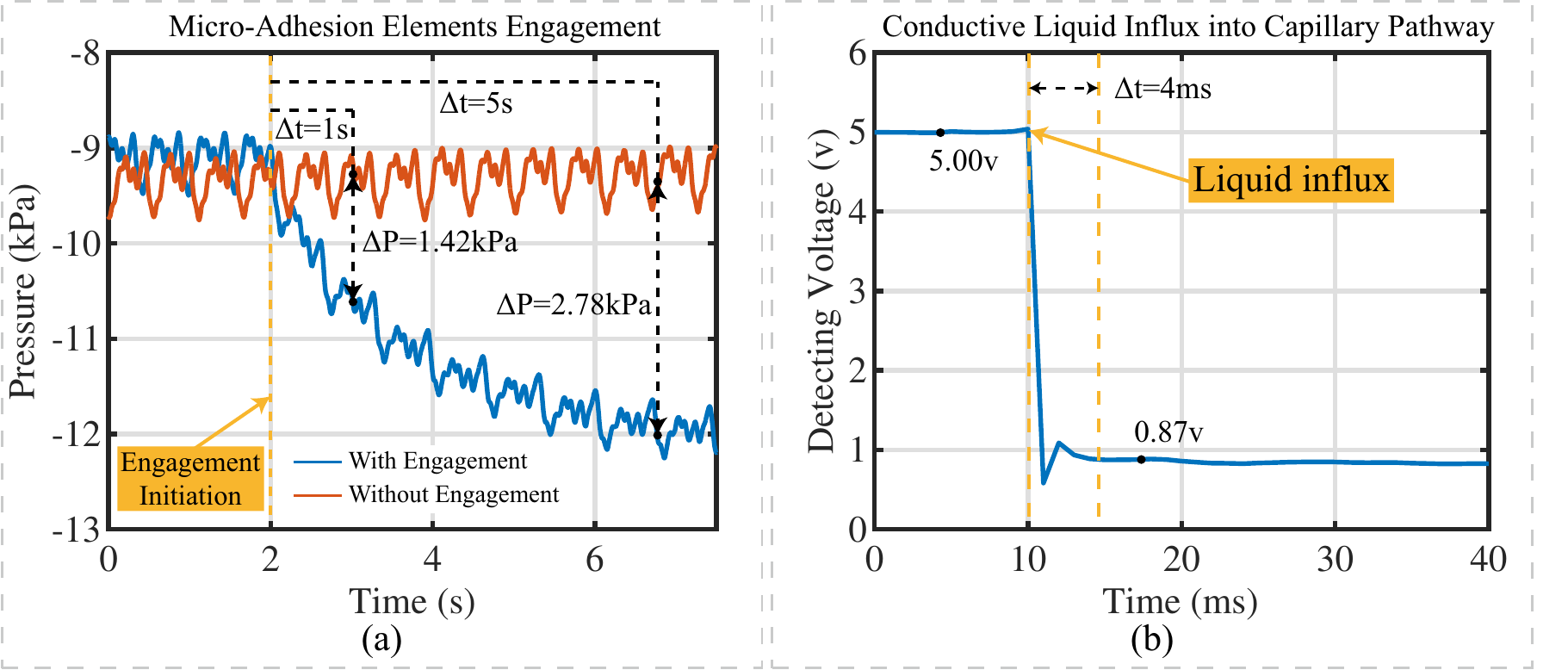}
\caption{Experimental results demonstrating smart contact detection and liquid ingress monitoring of micro-adhesion elements: (a) Pressure change following micro-adhesion element engagement; (b) Electric signal change when conductive liquid enters the capillary pathway.}
\label{figure_Pressure_Change_1110}
\end{figure}

\begin{figure}[htbp]
\centering
\includegraphics[width=0.9\linewidth]{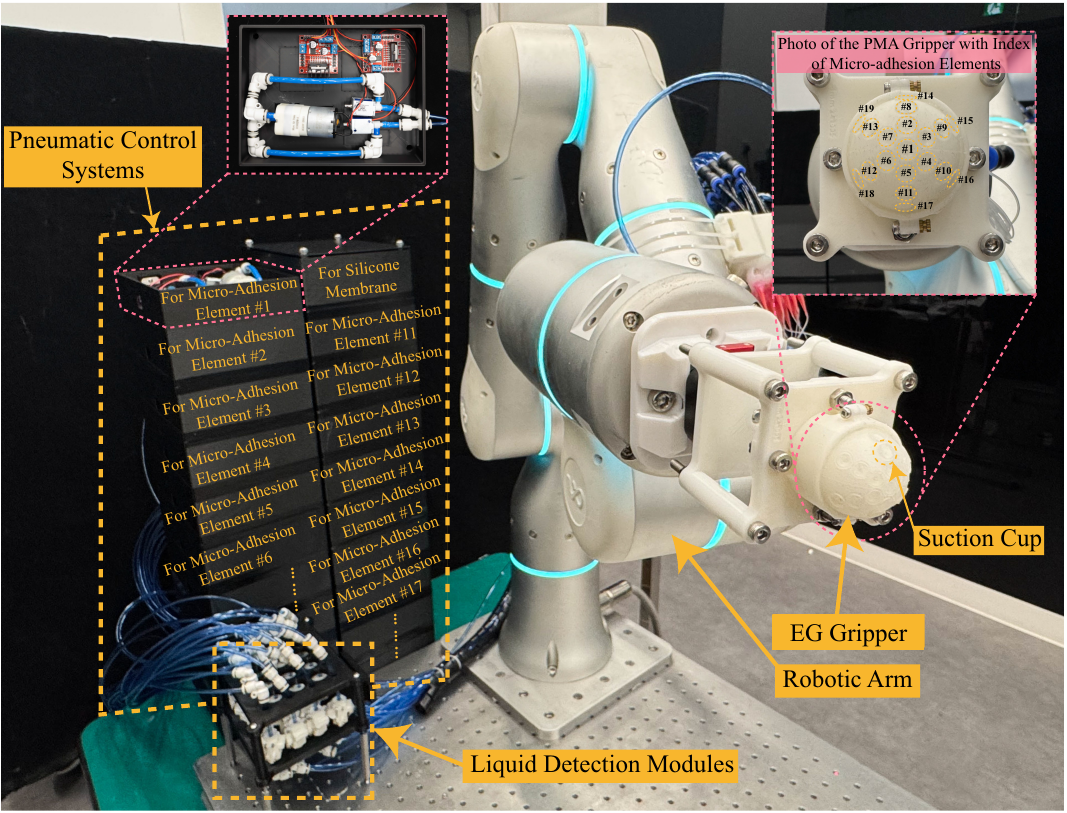}
\caption{Experimental setup demonstrating the capabilities of the EG gripper in cross-scale and cross-state object manipulation.}
\label{figure_demo_setup}
\end{figure}

\begin{figure}[htbp]
\centering
\includegraphics[width=1\linewidth]{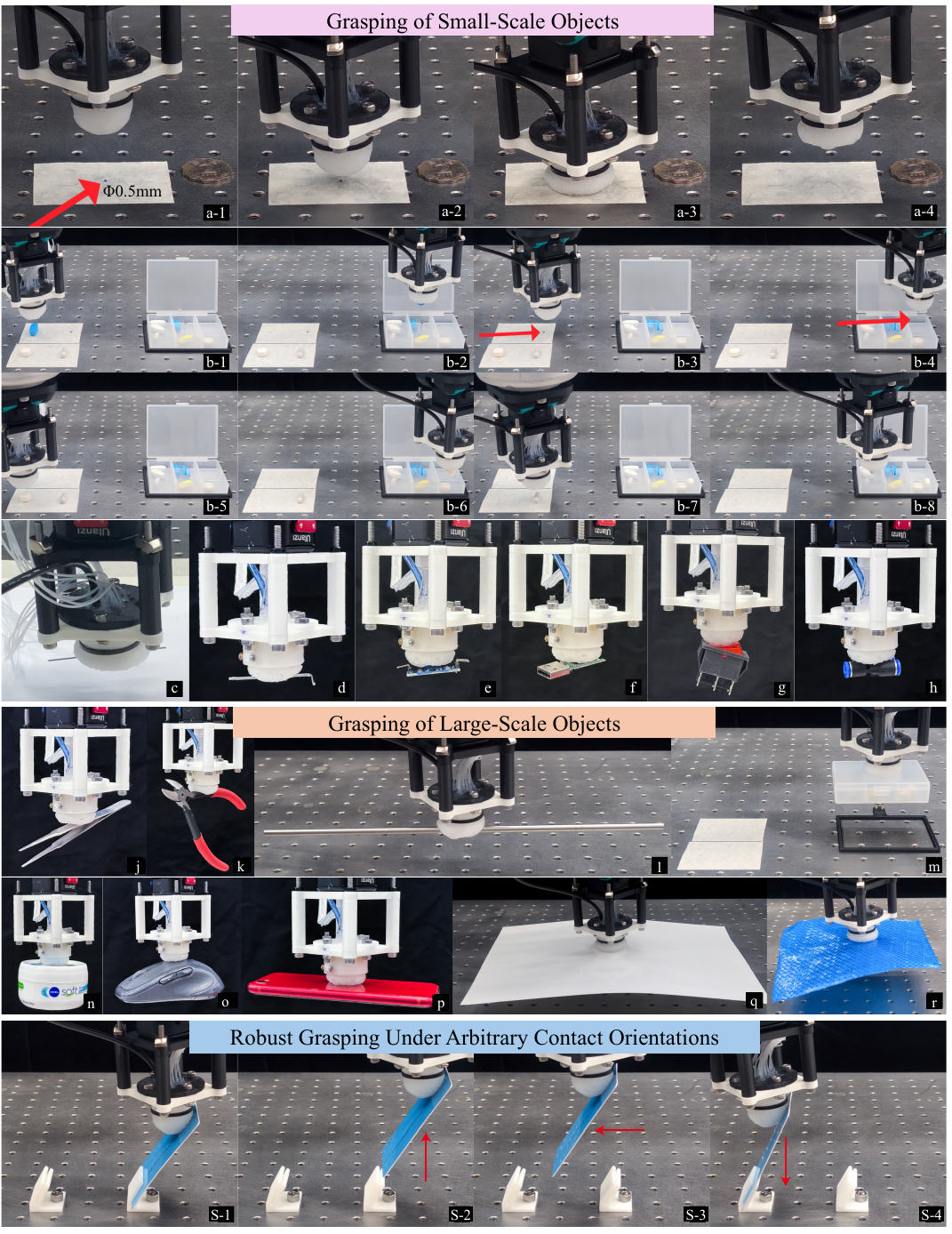}
\caption{Demonstration of the EG gripper's cross-scale grasping and orientation robustness. 
(a1–a4) Grasping of a sub-millimeter object ( $\phi$0.5 mm EPS particle). 
(b1–b8) Reliable handling of small pharmaceutical items with varying shapes and packaging. 
(c–h) Manipulation of everyday small-scale objects such as coins, SIM cards, batteries, and buttons. 
(i–r) Grasping of large-scale items (e.g., tools, bags, documents, and deformable packages) exceeding the gripper's contact base. 
(s1–s4) Stable grasping performance under varied contact angles, demonstrating tolerance to arbitrary approach orientations.}
\label{figure_demo_small}
\end{figure}

\subsection*{C. Demonstration of the Capability of the EG Gripper in Cross-Scale and Cross-State Object Manipulation}

In this section, the capabilities of the EG gripper in cross-scale and cross-state object manipulation are explored. The experimental setup for object manipulation is shown in Fig.~7. The EG gripper is mounted on a robotic arm (Rizon 4, Flexiv, China) through an adaptor. The liquid detection modules are placed on the side of the operation table, together with the pneumatic control systems. The robotic arm is programmed to allow the EG gripper to attach to the grasping object and grasp it.

The EG gripper’s cross-scale manipulation capability is demonstrated in Fig.~8(a–q), encompassing objects significantly smaller and larger than its own contact area. 

Fig.~8(a–h) illustrates grasping of small-scale targets. The gripper captures a 0.5\,mm-diameter EPS bead via the central DSE (Fig.~8a), showcasing capillary-only suction at submillimeter scale. In a pharmaceutical sorting task (Fig.~8b), it handles pills ranging from 1\,mm particles to 10\,mm capsules with varying shapes. Fig.~8(c–d) show stable grasping of a pencil lead and a small-diameter hex key, both narrower than the suction cup surface. Fig.~8(e–g) demonstrate handling of compact electronic modules with irregular or concave topographies, including a DC-DC converter, USB module, and rocker switch. Fig.~8(h) presents a PU tube connector with an M-shaped surface. These results confirm the gripper’s adaptability to fragile, narrow, and geometrically complex small objects. Fig.~8(i–q) highlights grasping of objects larger than the gripper. Fig.~8(i–k) demonstrate lifting of elongated rigid tools such as tweezers, pliers, and a steel rod, all exceeding the gripper’s footprint. Fig.~8(l–o) show manipulation of common items, including a pillbox, hand cream jar, convex-surfaced mouse, and a smartphone. Fig.~8(p–q) display handling of thin and deformable sheet-like structures, including an A4 paper and a flexible woven bag. These examples validate the EG gripper’s ability to conform to and lift soft, large-scale, or low-stiffness objects beyond its own structural size.

\begin{figure}[t]

\centering
\includegraphics[width=1\linewidth]{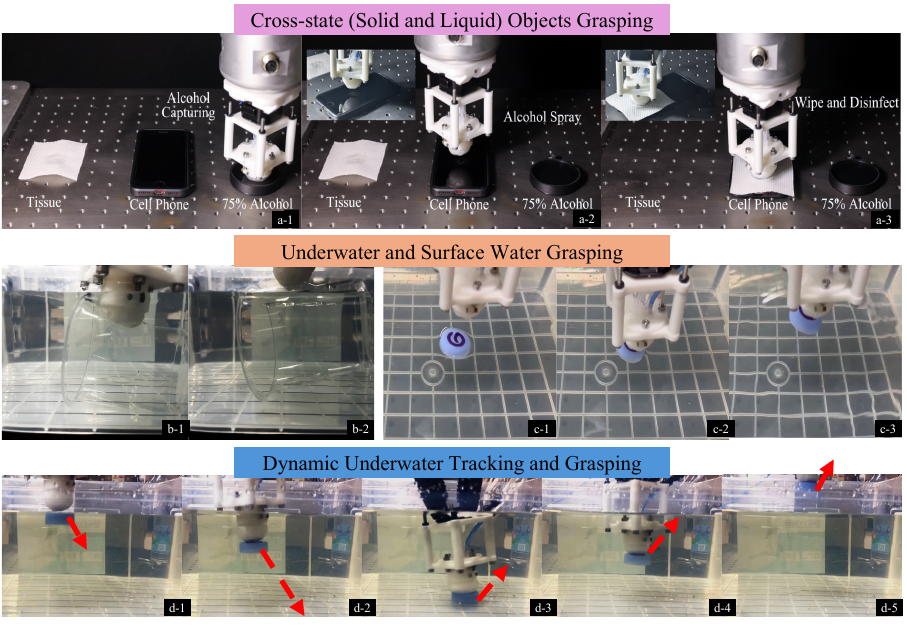}
\caption{Demonstration of the EG gripper's cross-state and underwater capabilities. 
(a1–a3) Sequential manipulation involving both solid and liquid targets (e.g., capturing alcohol spray and disinfecting a smartphone). 
(b1–b2) Underwater grasping of a slippery, curved-surface glass container.
(c1–c3) Surface water grasping of floating objects. 
(d1–d5) Dynamic underwater tracking and grasping of a moving target, highlighting robustness under fluid-induced disturbances.}
\label{figure_demo_liquid}
\end{figure}

The EG gripper’s capabilities in cross-state and underwater manipulation are shown in Fig.~9. In Fig.~9(a1–a3), the gripper performs a disinfection task involving both liquid and solid states: it captures alcohol via the central DSE, ejects it onto a smartphone screen, and subsequently grasps a tissue to wipe the surface. Fig.~9(b1–b2) illustrates underwater grasping of a smooth, curved glass container, demonstrating robust suction and lifting under submerged conditions. Fig.~9(c1–c3) shows collection of floating items on a water surface via surface suction without submerging, while Fig.~9(d1–d5) presents dynamic underwater tracking and grasping of a moving object, highlighting real-time adaptability and stable contact under motion and fluid-induced disturbances.

Together, these demonstrations validate the EG gripper’s scalability, compliance, and multimodal versatility in manipulating both liquids and solids across diverse sizes, surfaces, and environments.

The accompanying video highlights the EG gripper’s excellent dynamic performance in handling a wide range of cross-scale and cross-state grasping tasks, proving it to be one of the first systems in the field to achieve such integrated functionality.

\section*{V. Conclusion and Future Work}

We have presented the Everything-Grasping (EG) gripper, a robotic gripper capable of seamless cross-scale and cross-state manipulation using a unified, tactile sensing integrated platform. By synergistically combining distributed surface suction with internal granular jamming, the EG gripper achieves robust, single-mode grasping without requiring airtight sealing or material-specific tuning.

The gripper demonstrates reliable manipulation of objects ranging in surface area from 0.2\,mm$^2$ (e.g., glass beads) to over 62,000\,mm$^2$ (e.g., A4 sheets and woven bags)—spanning nearly five orders of magnitude. This unprecedented range exceeds the operational envelope of previously reported suction–grasping hybrids. Guided by the Tactile-Inferred Grasping Mode Selection (TIGMS) strategy, the system autonomously infers object state and size from distributed pressure and voltage feedback, enabling real-time, vision-free grasping across both solid and liquid targets.

Future work will address several remaining limitations and extensions. First, we will characterize long-term durability and stability of suction and jamming performance under repetitive, dynamic, and wet environments. Second, we aim to generalize the TIGMS framework to accommodate irregular-shaped or multi-contact objects by integrating shape-aware learning or contact clustering. Third, we plan to investigate extensions beyond conductive liquids by adapting the liquid detection module for broader material compatibility. Lastly, we will develop closed-loop control strategies leveraging embedded sensing and distributed force estimation to enable fully autonomous operation in cluttered, unstructured, or mobile scenarios such as drone-based or wearable applications.

\section*{Supplementary Material: Fabrication Process of the EG Gripper}
The fabrication process of the EG gripper is illustrated in Fig.~S1 and comprises four steps. The first step involves the fabrication of the silicone membrane. A set of molds (left mold, right mold, and bottom mold) was designed and 3D printed with resin material (Ledo 6060, Royal DSM, Netherlands). Thin steel rods, each with a length of 20~mm and a diameter of 0.6~mm, were inserted into holes in the molds (as shown in Fig.~S1) to create central apertures in the bowl-shaped suction cups for capillaries. Part A and Part B of liquid silicone (Ecoflex 00-30, Smooth-On, USA) were mixed in a 1:1 weight ratio and degassed using a vacuum tank. A syringe was used to inject the mixed silicone into the assembled mold through the vent in the center. In the second step, silicone capillaries (with an inner diameter of 0.3~mm and an outer diameter of 0.8~mm) were connected to the central apertures of the bowl-shaped suction cups using silicone adhesive (Sil-Poxy, Smooth-On, USA). In the third step, the EPS particles were filled into the silicone membrane. The silicone membrane was then sealed using 3D printed rigid caps and silicone adhesive. In the final step, the adapter for the robotic arm was installed on the top of the EG gripper to enable the gripper to connect to the robotic arm.

\begin{figure}[ht]
\centering
\renewcommand{\thefigure}{S1}  
\includegraphics[width=1\linewidth]{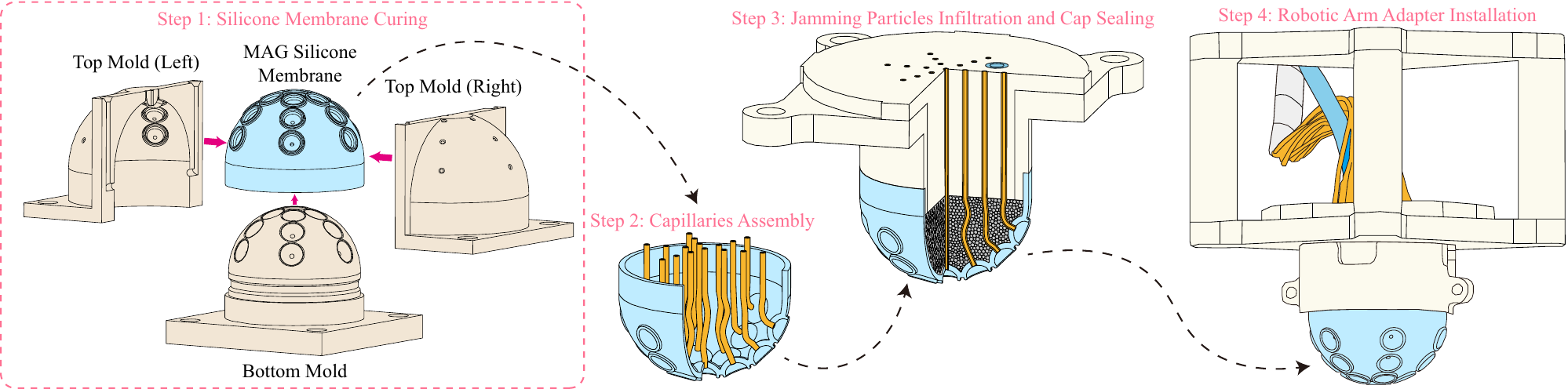}
\caption{Fabrication process of the EG gripper.}
\label{FIG_Fabrication}
\addtocounter{figure}{-1} 
\end{figure}

\section*{Acknowledgment}
This work was supported in part by the National Natural Science Foundation of China under Grant 52205032, in part by the Shun Hing Institute of Advanced Engineering, The Chinese University of Hong Kong, in part by Research Grants Council of Hong Kong (Ref. GRF No.
14204423 and 14207423), and in part by Guangdong Basic and Applied Basic Research Foundation under Grant 2023A1515010062

\bibliographystyle{unsrt}
\bibliography{references}

\begin{thebibliography}{10}

\bibitem{billard2019trends}
Aude Billard and Danica Kragic.
\newblock Trends and challenges in robot manipulation.
\newblock {\em Science}, 364(6446):eaat8414, 2019.

\bibitem{bicchi2000robotic}
Antonio Bicchi and Vijay Kumar.
\newblock Robotic grasping and contact: A review.
\newblock In {\em Proceedings 2000 ICRA. Millennium conference. IEEE international conference on robotics and automation. Symposia proceedings (Cat. No. 00CH37065)}, volume~1, pages 348--353. IEEE, 2000.

\bibitem{Prattichizzo2016}
Domenico Prattichizzo and Jeffrey~C. Trinkle.
\newblock {\em Grasping}, pages 955--988.
\newblock Springer International Publishing, Cham, 2016.

\bibitem{tai2016state}
Kevin Tai, Abdul-Rahman El-Sayed, Mohammadali Shahriari, Mohammad Biglarbegian, and Shohel Mahmud.
\newblock State of the art robotic grippers and applications.
\newblock {\em Robotics}, 5(2):11, 2016.

\bibitem{rus2015design}
Daniela Rus and Michael~T. Tolley.
\newblock Design, fabrication and control of soft robots.
\newblock {\em Nature}, 521(7553):467--475, May 2015.

\bibitem{wang2018toward}
Hongbo Wang, Massimo Totaro, and Lucia Beccai.
\newblock Toward perceptive soft robots: Progress and challenges.
\newblock {\em Advanced Science}, 5(9):1800541, 2018.

\bibitem{shintake2018soft}
Jun Shintake, Vito Cacucciolo, Dario Floreano, and Herbert Shea.
\newblock Soft robotic grippers.
\newblock {\em Advanced materials}, 30(29):1707035, 2018.

\bibitem{cianchetti2018biomedical}
Matteo Cianchetti, Cecilia Laschi, Arianna Menciassi, and Paolo Dario.
\newblock Biomedical applications of soft robotics.
\newblock {\em Nature Reviews Materials}, 3(6):143--153, 2018.

\bibitem{kim2013soft}
Sangbae Kim, Cecilia Laschi, and Barry Trimmer.
\newblock Soft robotics: a bioinspired evolution in robotics.
\newblock {\em Trends in Biotechnology}, 31(5):287--294, 2013.

\bibitem{kim2020untethered}
D.~I. Kim, S.~Song, S.~Jang, G.~Kim, J.~Lee, Y.~Lee, and S.~Park.
\newblock Untethered gripper-type hydrogel millirobot actuated by electric field and magnetic field.
\newblock {\em Smart Materials and Structures}, 29(8):085024, 2020.

\bibitem{yarali2022magneto}
E.~Yarali, M.~Baniasadi, A.~Zolfagharian, M.~Chavoshi, F.~Arefi, M.~Hossain, and M.~Bodaghi.
\newblock Magneto‐/electro‐responsive polymers toward manufacturing, characterization, and biomedical/soft robotic applications.
\newblock {\em Applied Materials Today}, 26:101306, 2022.

\bibitem{zhou2017soft}
Jianshu. Zhou, Shu. Chen, and Zheng. Wang.
\newblock A soft-robotic gripper with enhanced object adaptation and grasping reliability.
\newblock {\em IEEE Robotics and Automation Letters}, 2(4):2287--2293, 2017.

\bibitem{zhou2019soft}
Jianshu. Zhou, Xiaojiao. Chen, Ukyoung. Chang, Jui.~Ting. Lu, Clarisse. Ching.~Yau. Leung, Yonghua. Chen, and Zheng. Wang.
\newblock A soft-robotic approach to anthropomorphic robotic hand dexterity.
\newblock {\em IEEE Access}, 7:101483--101495, 2019.

\bibitem{gu2021analytical}
Guoying Gu, Dongxu Wang, Liang Ge, and Xinyu Zhu.
\newblock Analytical modeling and design of generalized pneu-net soft actuators with three-dimensional deformations.
\newblock {\em Soft Robotics}, 8(4):462--477, 2021.

\bibitem{zhou2023antagonistic}
Jianshu Zhou, Wei Chen, Hanwen Cao, Qiguang He, and Yunhui Liu.
\newblock Antagonistic pump with multiple pumping modes for on-demand soft robot actuation and control.
\newblock {\em IEEE/ASME Transactions on Mechatronics}, 2023.

\bibitem{crooks2016fin}
Wesley Crooks, Gregory Vukasin, Maureen O’Sullivan, William Messner, and Christopher Rogers.
\newblock Fin ray{\textregistered} effect inspired soft robotic gripper: From the robosoft grand challenge toward optimization.
\newblock {\em Frontiers in Robotics and AI}, 3:70, 2016.

\bibitem{shin2021universal}
Jin~Hyuk Shin, Jin~Gon Park, Dong~Il Kim, and Hyung~Seok Yoon.
\newblock A universal soft gripper with the optimized fin ray finger.
\newblock {\em International Journal of Precision Engineering and Manufacturing-Green Technology}, 8:889--899, 2021.

\bibitem{zhou2024dexterous}
Jianshu Zhou, Junda Huang, Qi~Dou, Pieter Abbeel, and Yunhui Liu.
\newblock A dexterous and compliant (dexco) hand based on soft hydraulic actuation for human inspired fine in-hand manipulation.
\newblock {\em IEEE Transactions on Robotics}, 2024.

\bibitem{zhou2020adaptive}
Jianshu Zhou, Yonghua Chen, Yong Hu, Zheng Wang, Yunquan Li, Guoying Gu, and Yunhui Liu.
\newblock Adaptive variable stiffness particle phalange for robust and durable robotic grasping.
\newblock {\em Soft robotics}, 7(6):743--757, 2020.

\bibitem{sinatra2019ultragentle}
Nicholas~R Sinatra, Caitlyn~B Teeple, Daniel~M Vogt, Kevin~K Parker, David~F Gruber, and Robert~J Wood.
\newblock Ultrag entle manipulation of delicate structures using a soft robotic gripper.
\newblock {\em Science Robotics}, 4(33):eaax5425, 2019.

\bibitem{yang2024multi}
C.~Yang, I.~D. Walker, D.~T. Branson, J.~S. Dai, T.~Sun, and R.~Kang.
\newblock A multi-tentacle gripper for dynamic capture.
\newblock {\em IEEE Transactions on Robotics}, 2024.
\newblock Early Access.

\bibitem{nguyen2023granulartendon}
Van~Phu Nguyen, Sunil~Bohra Dhyan, Boon~Siew Han, and Wai~Tuck Chow.
\newblock Universally grasping objects with granular–tendon finger: Principle and design.
\newblock {\em Micromachines}, 14(7):1471, 2023.

\bibitem{amend2012positive}
James~R Amend, Eric Brown, Nick Rodenberg, Heinrich~M Jaeger, and Hod Lipson.
\newblock A positive pressure universal gripper based on the jamming of granular material.
\newblock {\em IEEE Transactions on Robotics}, 28(2):341--350, 2012.

\bibitem{nguyen2023universally}
Van~Phu Nguyen, S.~B. Dhyan, B.~S. Han, and W.~T. Chow.
\newblock Universally grasping objects with granular—tendon finger: Principle and design.
\newblock {\em Micromachines}, 14(7):1471, 2023.

\bibitem{li2017passive}
Yingtian Li, Yonghua Chen, Yang Yang, and Ying Wei.
\newblock Passive particle jamming and its stiffening of soft robotic grippers.
\newblock {\em IEEE Transactions on robotics}, 33(2):446--455, 2017.

\bibitem{li2019vacuum}
Shuguang Li, Jacob~J. Stampfli, Hanjun~J. Xu, Ethan Malkin, Elias~V. Diaz, Daniela Rus, and Robert~J. Wood.
\newblock A vacuum-driven origami “magic-ball” soft gripper.
\newblock In {\em 2019 International Conference on Robotics and Automation (ICRA)}, pages 7401--7408. IEEE, 2019.

\bibitem{cao2024design}
Hanwen Cao, Jianshu Zhou, Kaix Chen, Qiguang He, Qi~Dou, and Yunhui Liu.
\newblock Design and optimization of an origami gripper for versatile grasping and manipulation.
\newblock {\em Advanced Intelligent Systems}, 6(12):2400271, 2024.

\bibitem{kim2021fourD}
Keumbee Kim, Yuanhang Guo, Jaehee Bae, Subi Choi, Hyeong~Yong Song, Sungmin Park, Kyu Hyun, and Suk-Kyun Ahn.
\newblock 4d printing of hygroscopic liquid crystal elastomer actuators.
\newblock {\em Small}, 17(23):2100910, 2021.

\bibitem{koivikko2021magnetically}
Antti Koivikko, Dirk~M. Drotlef, Metin Sitti, et~al.
\newblock Magnetically switchable soft suction grippers.
\newblock {\em Extreme Mechanics Letters}, 44:101263, 2021.

\bibitem{zhou2024design}
Jianshu Zhou, Junda Huang, Xin Ma, Andy Lee, Kazuhiro Kosuge, and Yun-Hui Liu.
\newblock Design, modeling, and control of soft syringes enabling two pumping modes for pneumatic robot applications.
\newblock {\em IEEE/ASME Transactions on Mechatronics}, 2024.

\bibitem{zhou2022bioinspired}
Jianshu Zhou, Hanwen Cao, Wei Chen, Shing~Shin Cheng, and Yun-Hui Liu.
\newblock Bioinspired soft wrist based on multicable jamming with hybrid motion and stiffness control for dexterous manipulation.
\newblock {\em IEEE/ASME Transactions on Mechatronics}, 28(3):1256--1267, 2022.

\bibitem{day2013microwedge}
Paul Day, Eric~V. Eason, Noe Esparza, David Christensen, and Mark Cutkosky.
\newblock Microwedge machining for the manufacture of directional dry adhesives.
\newblock {\em Journal of Micro and Nano-Manufacturing}, 1(1):011001, 2013.

\bibitem{hawkes2016three}
Elliot~W. Hawkes, Hao Jiang, and Mark~R. Cutkosky.
\newblock Three-dimensional dynamic surface grasping with dry adhesion.
\newblock {\em The International Journal of Robotics Research}, 35(8):943--958, 2016.

\bibitem{nguyen2018soft}
Phuoc~Van Nguyen, Nhu~Van Huynh, Tien-Thinh Phan, and Van-An Ho.
\newblock Soft grasping with wet adhesion: Preliminary evaluation.
\newblock In {\em 2018 IEEE International Conference on Soft Robotics (RoboSoft)}, pages 418--423. IEEE, 2018.

\bibitem{nguyen2019grasping}
Phuoc~Van Nguyen and Van~Anh Ho.
\newblock Grasping interface with wet adhesion and patterned morphology: Case of thin shell.
\newblock {\em IEEE Robotics and Automation Letters}, 4(2):792--799, 2019.

\bibitem{cacucciolo2019delicate}
Vito Cacucciolo, Jun Shintake, and Herbert Shea.
\newblock Delicate yet strong: Characterizing the electro-adhesion lifting force with a soft gripper.
\newblock In {\em 2019 2nd IEEE International Conference on Soft Robotics (RoboSoft)}, pages 108--113. IEEE, 2019.

\bibitem{gu2018soft}
Guoying Gu, Jiang Zou, Ruike Zhao, Xuanhe Zhao, and Xiangyang Zhu.
\newblock Soft wall-climbing robots.
\newblock {\em Science Robotics}, 3(25):eaat2874, 2018.

\bibitem{li2022bioinspired}
Jing Li, Zhenzhen Song, Chuandong Ma, Tonghang Sui, Peng Yi, and Jianlin Liu.
\newblock A bioinspired adhesive sucker with both suction and adhesion mechanisms for three-dimensional surfaces.
\newblock {\em Journal of Bionic Engineering}, 19(6):1671--1683, 2022.

\bibitem{bamotra2018fabrication}
Abhishek Bamotra, Pushpinder Walia, Avataram~Venkatavaradan Prituja, and Hongliang Ren.
\newblock Fabrication and characterization of novel soft compliant robotic end-effectors with negative pressure and mechanical advantages.
\newblock In {\em 2018 3rd International Conference on Advanced Robotics and Mechatronics (ICARM)}, pages 369--374. IEEE, 2018.

\bibitem{koivikko20213d}
Anastasia Koivikko, Dirk-Michael Drotlef, Cem~Balda Dayan, Veikko Sariola, and Metin Sitti.
\newblock 3d-printed pneumatically controlled soft suction cups for gripping fragile, small, and rough objects.
\newblock {\em Advanced Intelligent Systems}, 3(9):2100034, 2021.

\bibitem{pham2019critically}
Hung Pham and Quang-Cuong Pham.
\newblock Critically fast pick-and-place with suction cups.
\newblock In {\em 2019 International Conference on Robotics and Automation (ICRA)}, pages 3045--3051. IEEE, 2019.

\bibitem{liu2023hybrid}
Fukang Liu, Fuchun Sun, Bin Fang, Xiang Li, Songyu Sun, and Huaping Liu.
\newblock Hybrid robotic grasping with a soft multimodal gripper and a deep multistage learning scheme.
\newblock {\em IEEE Transactions on Robotics}, 39(3):2379--2399, 2023.

\bibitem{wang2020dual}
Zhongkui Wang, Keung Or, and Shinichi Hirai.
\newblock A dual-mode soft gripper for food packaging.
\newblock {\em Robotics and Autonomous Systems}, 125:103427, 2020.

\bibitem{wu2022glowing}
Mingxin Wu, Xingwen Zheng, Ruosi Liu, Ningzhe Hou, Waqar~Hussain Afridi, Rahdar~Hussain Afridi, Xin Guo, Jianing Wu, Chen Wang, and Guangming Xie.
\newblock Glowing sucker octopus (stauroteuthis syrtensis)‐inspired soft robotic gripper for underwater self‐adaptive grasping and sensing.
\newblock {\em Advanced Science}, 9(17):2104382, 2022.

\bibitem{washio2022design}
Shogo Washio, Kieran Gilday, and Fumiya Iida.
\newblock Design and control of a multi-modal soft gripper inspired by elephant fingers.
\newblock In {\em 2022 IEEE/RSJ International Conference on Intelligent Robots and Systems (IROS)}, pages 4228--4235. IEEE, 2022.

\bibitem{pan2024transformable}
Tianle Pan, Jianshu Zhou, Zihao Zhang, Huayu Zhang, Jinfei Hu, Jiajun An, Yunhui Liu, and Xin Ma.
\newblock Transformable soft gripper: Uniting grasping and suction for amphibious cross-scale objects grasping.
\newblock {\em Soft Robotics}, 2024.

\bibitem{chathuranga2013biomimetic}
Damith~Suresh Chathuranga, Zhongkui Wang, Van~Anh Ho, Atsushi Mitani, and Shinichi Hirai.
\newblock A biomimetic soft fingertip applicable to haptic feedback systems for texture identification.
\newblock In {\em 2013 IEEE International Symposium on Haptic Audio Visual Environments and Games (HAVE)}, pages 29--33, Dalian, China, October 2013. IEEE.

\bibitem{qi2022sea}
Qiukai Qi, Chaoqun Xiang, Van~Anh Ho, and Jonathan Rossiter.
\newblock A sea-anemone-inspired, multifunctional, bistable gripper.
\newblock {\em Soft Robotics}, 9(6):1040--1051, 2022.

\bibitem{cui2021enhancing}
Yafeng Cui, Xin-Jun Liu, Xuguang Dong, Jingyi Zhou, and Huichan Zhao.
\newblock Enhancing the universality of a pneumatic gripper via continuously adjustable initial grasp postures.
\newblock {\em IEEE Transactions on Robotics}, 37(5):1604--1618, 2021.

\bibitem{brown2010universal}
Eric Brown, Nicholas Rodenberg, John Amend, Annan Mozeika, Erik Steltz, Mitchell~R Zakin, Hod Lipson, and Heinrich~M Jaeger.
\newblock Universal robotic gripper based on the jamming of granular material.
\newblock {\em Proceedings of the National Academy of Sciences}, 107(44):18809--18814, 2010.

\end{thebibliography}

\end{document}